\documentclass[runningheads,orivec]{llncs}
\usepackage[T1]{fontenc}
\usepackage[dvipsnames]{xcolor}
\usepackage{graphicx}
\usepackage{tabularx}
\usepackage{amsmath}
\usepackage{amssymb}
\usepackage{mathtools}
\usepackage{todonotes}
\usepackage{bm}
\usepackage{float}
\usepackage{hyperref}
\usepackage[export]{adjustbox}
\usepackage[misc]{ifsym}

\newcommand{\R}{\mathbb{R}}

\newcommand{\vx}{\bm{x}}
\newcommand{\vy}{\bm{y}}
\newcommand{\vz}{\bm{z}}

\newcommand{\rx}{\mathbf{x}}
\newcommand{\ry}{\mathbf{y}}

\begin{document}
\title{Synthesizing Accurate and Realistic T1-weighted Contrast-Enhanced MR Images using Posterior-Mean Rectified Flow}
\titlerunning{Posterior-Mean Rectified Flow for Realistic, Accurate Virtual Contrast MRI}

\author{Bastian Brandst\"otter$^{1\ \text{0009-0002-3752-3051}}$ \and Erich Kobler$^{1,2,3\ \text{(\Letter)}\ \text{0000-0001-5167-4804}}$ }
\authorrunning{B. Brandst\"otter, E. Kobler}
\institute{$^1$Institute for Machine Learning, Johannes Kepler University Linz, Austria\\
\texttt{\{bastian.brandstoetter,erich.kobler\}@jku.at}\\
$^2$LIT AI Lab, Johannes Kepler University Linz, Austria\\
$^3$Department of Virtual Morphology, Johannes Kepler University Linz, Austria}

\maketitle              

\bibliographystyle{splncs04}

\begin{abstract}
Contrast-enhanced (CE) T1-weighted MRI is central to neuro-oncologic diagnosis but requires gadolinium-based agents, which add cost and scan time, raise environmental concerns, and may pose risks to patients. In this work, we propose a two-stage Posterior-Mean Rectified Flow (PMRF) pipeline for synthesizing volumetric CE brain MRI from non‐contrast inputs. First, a patch-based 3D U-Net predicts the voxel-wise posterior mean (minimizing MSE). Then, this initial estimate is refined by a time‐conditioned 3D rectified flow to incorporate realistic textures without compromising structural fidelity.
We train this model on a multi-institutional collection of paired pre- and post-contrast T1w volumes (BraTS 2023-2025).
On a held-out test set of 360 diverse volumes, our best refined outputs achieve an axial FID of $12.46$ and KID of $0.007$ ($\sim 68.7\%$ lower FID than the posterior mean) while maintaining low volumetric MSE of $0.057$ ($\sim 27\%$ higher than the posterior mean).
Qualitative comparisons confirm that our method restores lesion margins and vascular details realistically, effectively navigating the perception-distortion trade-off for clinical deployment.
\end{abstract}

\keywords{MRI synthesis \and Contrast enhancement \and Rectified flow \and Perception-distortion trade-off}

\section{Introduction}

Contrast-enhanced (CE) T1w MRI is invaluable for detecting and delineating lesions, but administering gadolinium-based contrast agents (GBCAs) has drawbacks -- it adds cost, lengthens exam time, increases pollution, and carries patient risks~\cite{Bendszus-2024-gbcm}.
This has motivated research into generating virtual CE MRI from non-CE scans, allowing us to reap its diagnostic benefits without injecting an exogenous agent.
Previous studies have explored deep learning models such as convolutional neural networks (CNNs) \cite{Calabrese-2021-cnn,Osman-2023-ddrc} or generative adversarial networks (GANs) \cite{Solak2024-gk,Zoghby2024-gk} to synthesize CE T1w images from pre-contrast inputs.
These have shown promising qualitative results, yet they often struggle to balance voxel-wise fidelity against perceptual realism.
Typically, a model optimized purely for MSE yields blurred but accurate images, whereas models aiming for more realism, typically using adversarial losses, produce sharper images that may deviate more from the true intensities.
This tension was formalized by Blau and Michaeli~\cite{Blau-2018-per-dist} as the perception-distortion trade-off: an estimator can either achieve low distortion (high fidelity to the ground truth) or high perceptual quality, but improving one inevitably worsens the other. In tumor imaging, both voxel-wise accuracy and perceptual realism are essential for diagnostic assessments.

In this work, we address this challenge using the Posterior-Mean Rectified Flow (PMRF) framework introduced by Ohayon et al.~\cite{ohayon2025posteriormean}.
PMRF introduces a principled two-stage approach: first, it predicts the posterior mean (i.e., the minimum-MSE estimator), and then applies a generative rectified flow to refine this estimate by injecting realistic details with minimal deviation.
We extend this paradigm to volumetric medical imaging and propose a novel pipeline for synthesizing CE T1w MRI from non-CE T1w MRI.
Our method explicitly decouples distortion minimization and perceptual enhancement into two separate stages, allowing for targeted optimization of each.
To manage memory constraints, we process overlapping 3-D patches independently and aggregate them using a Hann-windowed fusion strategy. This results in CE images that are quantitatively accurate and, crucially, perceptually closer to real CE scans when assessed by human observers.

\section{Methodology}

\begin{figure}
    \centering
    \includegraphics[width=\linewidth]{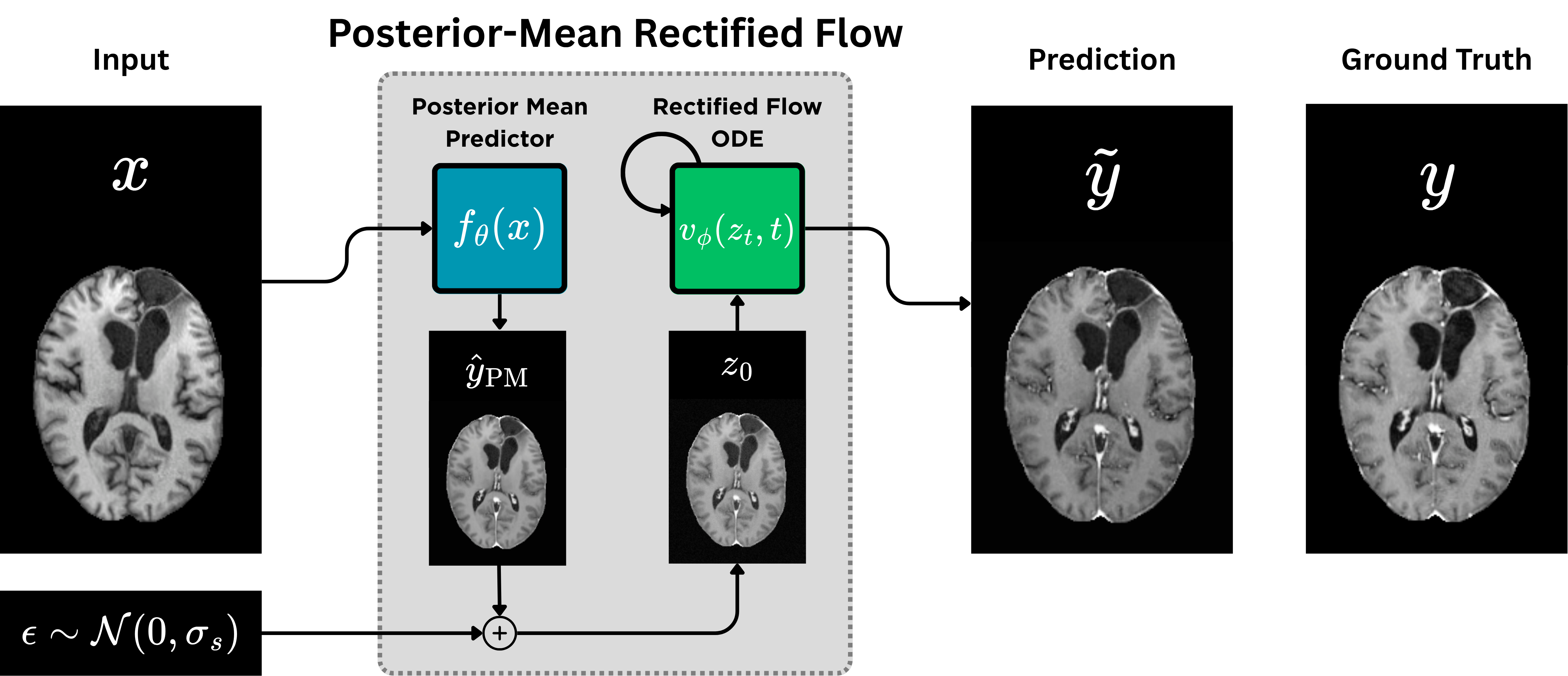}    
    \caption{Overview of the two-stage Posterior-Mean Rectified Flow (PMRF) pipeline. First, a model $f_\theta$ predicts the posterior mean estimate, which is slightly perturbed using Gaussian noise before being refined by a rectified flow (governed by the vector field $v_\phi$) synthesizing accurate and realistic contrast-enhanced T1-weighted (CE T1w) images.}
    \label{fig:pipeline}
\end{figure}

Our methodology consists of two sequential stages, as illustrated in Figure~\ref{fig:pipeline}. Let $\vx\in\mathbb{R}^{d}$ be a non-CE T1w image and $\vy\in\R^d$ its true contrast-enhanced counterpart, each realizations of a random vector $\rx$ and $\ry$, respectively.
Stage~1 trains a 3D U-Net $f_{\theta}$ to output the posterior mean $\hat \vy_{\text{PM}} = f_{\theta}(\vx)$ approximating the minimum-MSE estimator.
Stage~2 applies a rectified flow $v_{\phi}(\vz,t)$ that starts from a perturbed version of this estimate, $\vz_{0}$, and evolves it under the associated ordinary differential equation (ODE) yielding $\tilde \vy$, a refined volume that retains the structural accuracy of $\hat \vy_{\text{PM}}$ while adding realistic high-frequency detail.
By decoupling the objectives in this way, the pipeline can achieve a better fidelity-realism balance than a single model trying to do both~\cite{ohayon2025posteriormean}.
In the following, we detail each stage and the training procedure.

\subsubsection{Stage~1: Posterior-Mean Predictor (3D U-Net)}

A 3D residual U-Net~\cite{Khanna-2020-resunet} implements Stage~1 by taking a non-CE image \(\vx\) and predicting the corresponding CE image \(\vy\).
The network is trained to minimize the plain voxel-wise MSE, i.e.

\begin{equation}
\mathcal{L}_{\mathrm{MSE}}(\theta) = \mathbb{E}_{(\rx,\ry)\sim \mathcal{D}}\big[\,\|f_{\theta}(\rx) - \ry\|_2^2\,\big]\ ,
\end{equation}

where $\mathcal{D}$ denotes the distribution of the non-CE and CE image pairs.
As a result, $f_{\theta}$ learns to approximate the conditional posterior mean of the CE image given the non-CE input, meaning it yields the lowest possible average $\ell_2$ error.
In practice, this causes the prediction to be overly smoothed: the network confidently generates enhancement patterns that are consistently implied by the input, e.g.~a tumor’s solid enhancing core if the model has seen many similar cases, but it averages out ambiguous high-frequency details, e.g.~subtle, less-predictable enhancement at tumor margins or fine vascular structures.
The result is a high-fidelity yet somewhat blurred output.
We use a 3D U-Net architecture with residual blocks and skip connections, which is well-suited to capture both global context and fine local details in volumetric data.
To handle high-resolution volumes without exceeding memory limits, the U-Net is trained on randomly sampled $64^3$ patches.
During each training iteration, a random patch from a non-CE image is fed into the model being learned to predict corresponding patch from the registered CE image.
This patch-based training strategy not only makes training feasible on a single GPU, but also effectively augments the data by providing many random sub-volume examples per full scan.
The output of Stage~1 alone can serve as a virtual CE MRI with excellent quantitative fidelity (low MSE).
However, owing to the perception-distortion trade-off, this output lacks texture and subtle detail -- it often appears overly smooth or “denoised” compared to a real post-contrast scan (see Figure~\ref{fig:qualitative_examples}).
This motivates Stage~2, which will focus on adding back realistic details.

\subsubsection{Stage~2: Rectified Flow Refiner}
treats the first-stage posterior-mean volume as the starting point of an optimal-transport ODE that moves every voxel along a straight line toward the true contrast image, and it learns the governing vector field by flow-matching.

For any training pair \((\vx,\vy)\), let \(\hat\vy_{\text{PM}}\) denote the Stage~1 prediction.
We create the starting sample \(\vz_0\) by adding a small, i.i.d. Gaussian perturbation to the posterior mean:
\begin{equation}
\vz_0 = \hat\vy_{\mathrm{PM}} + \sigma_s \,\varepsilon,\qquad\varepsilon\sim\mathcal N(0,I),
\end{equation}
where $\sigma_s$ is a tunable diversity factor. This pushes \(\vz_0\) off the single posterior-mean point and lets the refiner cover the true conditional posterior support.

Following PMRF's optimal transport view \cite{ohayon2025posteriormean}, we define a deterministic coupling between the starting sample \(\vz_0\) and the target image \(\vy\) by the linear path
\begin{equation}
\vz_t = (1-t)\,\vz_0 + t\,\vy,\qquad 0\le t\le1 .
\end{equation}
Every voxel therefore travels on the unique straight segment that joins its initial value to its ground-truth value. Every voxel therefore travels on the unique straight segment that joins its initial value to its ground-truth value. This linear path admits a closed-form ODE solution and yields efficient simulation~\cite{Liu23-flow}. The path induces a constant vector field 
\begin{equation}
v^\star(\vz_t,t) = \dot\vz_t = \vy - \vz_0.
\end{equation}

We train a neural network \(v_\phi(\vz,t)\) implemented as a time-conditioned 3D U-Net to approximate this ground-truth field.
Training amounts to solving the initial-value problem
\begin{equation}
\dot{\vz}_t=v_\phi(\vz_t,t),
\end{equation}
with our previously established $\vz_0$ as the starting point at $t=0$. After unit time, the solution ideally reaches \(\vz_1=\vy\). The terminal point \(\vy\) is simply the corresponding ground-truth CE patch, so the learned flow transports samples from the slightly noisy posterior-mean distribution to the empirical distribution of real CE volumes.

Sampling a random time $t\sim\mathcal U[0,1]$ and evaluating the intermediate volume \(\vz_t\), the flow-matching loss regresses the network output onto the ideal constant vector field:
\begin{equation}
\mathcal L_{\mathrm{flow}}(\phi)
= \mathbb E_{(\rx,\ry),\,t}\!\left[\,
\|v_\phi(\vz_t,t)-(\vy-\vz_0)\|_2^2
\right].
\end{equation}
Unlike adversarial or score-matching objectives, this least-squares loss does not require simulation or internal ODE integration, directly reducing the quadratic transport cost at each training step.

At inference, we integrate the learned ODE from \(\vz_{0}\) using \(K\) explicit Euler steps:
\begin{equation}
\vz_{k+1} = \vz_k + \Delta t\, v_\phi(\vz_k, t_k),\quad \Delta t=\tfrac1K,\; t_k=\tfrac{k}{K},\; k=0,\dots,K-1.
\end{equation}
Because rectified flows learn nearly linear trajectories, high-quality results are obtained even with few steps, yielding a fast, single-pass refinement that adds realistic textures while preserving voxel-wise accuracy.

\subsection{Baseline Models}

We treat our Stage‑1 posterior‑mean predictor as an in‑study baseline: it is a residual 3‑D U‑Net closely resembling architectures already used for virtual CE T1w synthesis (e.g., \cite{Osman-2023-ddrc}). To probe the other extreme of the perception-distortion spectrum, we additionally trained a rectified flow model in the sense of Liu et al. \cite{Liu23-flow}. The rectified flow baseline is conditioned directly on the non-CE T1w image perturbed with additive Gaussian noise of standard deviation $\sigma_s$:
\[
\vz_0 = \vx + \sigma_s \varepsilon,\; \varepsilon\sim\mathcal N(0,I).
\]
Architecture, loss (flow-matching) and all hyper-parameters mirror Stage 2 of PMRF; only the conditioning differs.
We refer to this model as the \textbf{RF baseline} in this paper.

\subsection{Dataset and Preprocessing}
We gathered a total of 7,196 paired non-CE and CE T1w brain MRI volumes by integrating data from the BraTS 2023-2025 datasets~\cite{brats-2025,Karargyris2023-fedbench}.
These datasets include a variety of pathologies, such as high- and low-grade gliomas, metastases, meningiomas, and post-surgical cavities, from several different institutions.
The data was skull-stripped, and each scan was aligned to an isotropic resolution of $1\text{mm}^3$ by the BraTS challenge team.
Subsequently, we divided the entire dataset into 80\% for training, 15\% for validation, and 5\% as a held-out test set.
Each volume was $Z$-score normalized (separately for non-CE T1w and CE T1w) to reduce scanner biases.
No other augmentations were used for training beyond randomly sampling $64^3$ patches, which enabled large batch sizes and exposed the networks to all kinds of tissue.
At inference, both stages operate patch-wise with specified overlaps and Hann window blending~\cite{Pielawski-2020-hann} to seamlessly reconstruct full volumes.

\section{Experiments and Results}
In the following, we present a quantitative and qualitative evaluation of the proposed approach.
Further details such as data splits, training schedules, optimizer settings, patch sampling strategy, and the training/validation loss curves for both stages are provided in Appendix~\ref{appendix:experiment_config}.

\subsection{Quantitative Evaluation}

We assess \emph{perceptual quality} with Fréchet Inception Distance (FID) \cite{heusel-2017-fid} and Kernel Inception Distance (KID) \cite{BinkowskiSAG18-kid}, both computed on 2D axial slices.
To this end, we extract the 2048-dimensional \texttt{pool\_3} features of a pretrained Inception-V3 network for nine equally spaced axial slices (central slice $\pm\{0,10, 20, 30, 40\}$ voxels) and compare their distributions against the corresponding slice set from the training corpus.
Although InceptionV3 is pretrained on natural images, recent work demonstrates that deep feature distances computed with such out-of-domain encoders correlate strongly with radiologists’ assessments of MR image quality~\cite{Adamson-FeatDist-2025}, validating the use of FID and KID in this medical imaging context.
\emph{Distortion} is measured volumetrically with mean-squared-error (MSE) and peak-signal-to-noise ratio (PSNR), as well as mean slice-wise structural similarity (SSIM).

\begin{table}[h]
  \centering
  \caption{Quantitative comparison of CE T1w image synthesis methods on the held-out test set. Best overall values in each column are \textbf{bold}, second best are \underline{underlined}. For both RF and PMRF, we report the hyper-parameter pair (patch overlap $=32$, number of flow steps $K=200$) that yielded the best FID on the test set. }
\begin{tabular}{lcc|ccc|cc}
\hline
Method & Overlap & $K$ & MSE~$\downarrow$ & PSNR (dB)~$\uparrow$ & SSIM~$\uparrow$ & FID$_\text{axial}$~$\downarrow$ & KID$_\text{axial}$~$\downarrow$ \\
\hline
\hline
Residual U-Net & 32 & - & \textbf{0.0450} & \textbf{34.82} & \textbf{0.9524} & 39.76 & 0.0314 \\
RF baseline & 32 & 200 & 0.0623 & 33.30 & 0.9434 & \textbf{11.66} & \textbf{0.0068} \\
\textbf{PMRF (ours}) & 32 & 200 & \underline{0.0570} & \underline{33.64} & \underline{0.9442} & \underline{12.46} & \underline{0.0071} \\
\hline
\end{tabular}
  \label{tab:comparison}
\end{table}

In Table~\ref{tab:comparison}, we compare all perceptual quality and distortion measures between the residual U-Net, which is used simultaneously as a baseline and as the posterior mean predictor for PMRF, the rectified flow (RF) baseline, and our two-stage approach (PMRF).
The residual U-Net baseline delivers the lowest distortion (best MSE/PSNR/SSIM), but at the cost of poor perceptual quality (highest FID/KID).
Conversely, the RF baseline drives FID/KID to the extreme left of the perception-distortion plane, yet incurs the largest distortion (worst MSE/PSNR/SSIM), confirming that emphasizing perceptual realism alone can severely degrade pixel‑wise fidelity.
Our \textbf{PMRF} attains an FID of \textbf{less than a third} of the residual U-Net while increasing MSE by only \textbf{$\sim 27 \%$}, and outperforms the RF baseline on every distortion metric; PMRF therefore offers a favorable balance between realism and fidelity.

\begin{figure}[H]
  \centering
  \includegraphics[width=\textwidth]{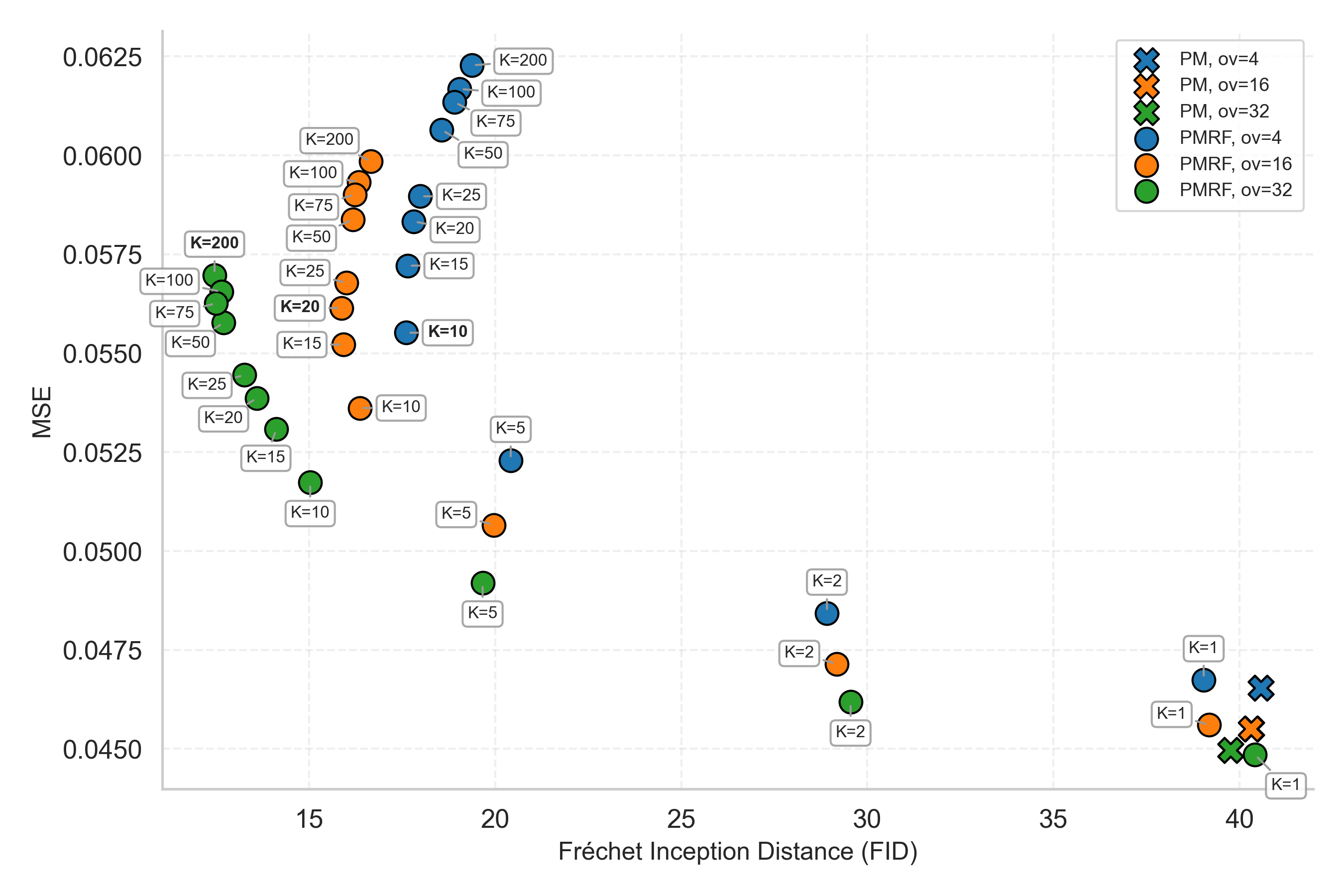}
  \caption{Perception-distortion trade-off.  Axial FID (lower is better) versus volumetric MSE (lower is better) for the posterior-mean baseline (PM) and the two-stage PMRF with Euler integration steps $K\in\{1,2,5,10,15,20,25,50,75,100,200\}$.  Marker colour denotes patch overlap during inference: \textcolor{blue}{4 voxels ($6.25\%$, blue)}, \textcolor{orange}{16 voxels ($25\%$, orange)}, and \textcolor{ForestGreen}{32 voxels ($50\%$, green)}. Each point from Stage~2 is annotated with its $K$ value.}
  \label{fig:axial_fid_vs_mse}
\end{figure}

Next, we evaluate how the number of flow steps~$K$ effects the results.
Fig.~\ref{fig:axial_fid_vs_mse} depicts the PM baseline along with PMRF results for different $K$ in the distortion (MSE) and perception (FID) plane for a different number of overlapping voxels. 
We identified the following key observations:
\begin{enumerate}
  \item \textbf{Overlap 4 voxels (blue).}  
        As $K$ increases from 1 to $\sim$10, FID falls sharply while MSE rises slightly, tracing the bottom branch of a sideways “U”.  Beyond $K=10$ the curve bends back: additional steps no longer improve perceptual quality and progressively increase distortion.
  \item \textbf{Overlap 16 voxels (orange).}  
        The medium-overlap setting inherits the same U-shaped trend but its entire curve is shifted to the left with respect to the 4-voxel case--indicating gains in perceptual fidelity. The minimum FID is reached at $K=20$.
  \item \textbf{Overlap 32 voxels (green).}  
        With the largest overlap, FID and MSE improve (or remain flat) monotonically up to our maximum tested setting of $K=200$; Consequently, the best operating point we were able to evaluate is $(\text{overlap}=32,;K=200)$.
  \item \textbf{Effect of a single refinement step ($K=1$).}  
        For overlap 32, the first flow step marginally \emph{improves} MSE while slightly \emph{worsening} FID; for overlap 4 and 16, the behaviour is inverted.
\end{enumerate}

\begin{table}[h]
  \centering
  \caption{Distortion and perception metrics for different combinations of overlap and number of Euler steps $K$ on the 360-volume test set. PM = Stage~1 posterior-mean prediction; PMRF = Stage~2 output after $K$ Euler steps. \textbf{Row selection}: all posterior-mean runs are listed; for the refined runs (PMRF) only the single configuration with the lowest FID is kept for each overlap, except that the PMRF model with 32-voxel overlap and $K{=}1$ is additionally retained because it achieves the global best MSE. Best overall values in each column are \textbf{bold}.}
\begin{tabular}{lcc|ccc|cc}
\hline
Model & Overlap & $K$ & MSE~$\downarrow$ & PSNR (dB)~$\uparrow$ & SSIM~$\uparrow$ & FID$_\text{axial}$~$\downarrow$ & KID$_\text{axial}$~$\downarrow$ \\
\hline
\hline
PM (Res U-Net) & 4 & - & 0.0465 & 34.66 & 0.9516 & 40.57 & 0.0326 \\
PM (Res U-Net) & 16 & - & 0.0455 & 34.75 & 0.9521 & 40.31 & 0.0321 \\
PM (Res U-Net) & 32 & - & 0.0450 & \textbf{34.82} & \textbf{0.9524} & 39.76 & 0.0314 \\
\hline
PMRF (ours) & 4 & 10 & 0.0555 & 33.76 & 0.9441 & 17.61 & 0.0122 \\
PMRF (ours) & 16 & 20 & 0.0561 & 33.70 & 0.9436 & 15.87 & 0.0104 \\
PMRF (ours) & 32 & 1 & \textbf{0.0449} & 34.81 & 0.9516 & 40.41 & 0.0328 \\
PMRF (ours) & 32 & 200 & 0.0570 & 33.64 & 0.9442 & \textbf{12.46} & \textbf{0.0071} \\
\hline
\end{tabular}
  \label{tab:metrics_summary}
\end{table}

Across all three overlap settings, the two-stage PMRF pipeline convincingly pushes operating points to the left in the perception-distortion plane relative to the Stage~1 PM. Increasing the patch overlap reduces initial distortion by providing the refiner with more spatial context, while increasing the number of flow steps \emph{K} mainly improves perceptual realism until an overlap-dependent optimum is reached. Higher patch overlap also leads to better maximum perceptual scores at the respective optimal flow steps. With the optimal configuration ${( \text{overlap}=32,\;K=200 )}$ we obtain a $\sim 68.7\%$ relative reduction in axial FID ($\sim 39.76 \to \sim 12.46$) for only a $\sim 27 \%$ relative increase in MSE ($\sim 0.045 \to \sim 0.057$), thereby navigating the perception-distortion trade-off much more effectively than the single-stage baseline.
An overview of evaluation results for different configurations of overlap and Euler steps for Stage~1 and Stage~2 can be examined in Table~\ref{tab:metrics_summary}.
For metrics measured on all evaluated Stage~1, RF baseline and Stage~2 inference configurations, refer to Table~\ref{tab:metrics_full_summary} in Appendix~\ref{appendix:full_evaluation}. 

\subsection{Qualitative Results}

\begin{figure}[H]
  \centering
  {\tiny
   \setlength{\tabcolsep}{0pt}%
   \begin{tabularx}{\textwidth}{@{}*{5}{>{\centering\arraybackslash}X}@{}}
    \includegraphics[width=.2\textwidth]{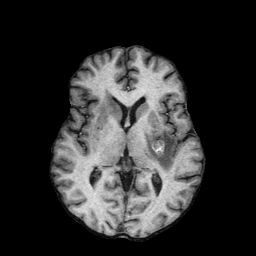}&%
    \includegraphics[width=.2\textwidth]{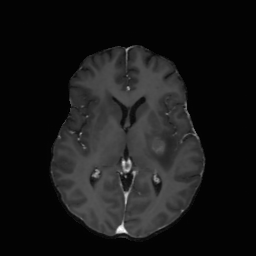}&%
    \includegraphics[width=.2\textwidth]{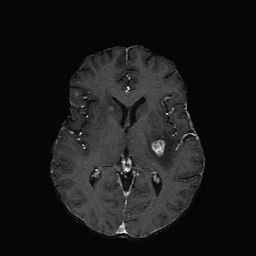}&%
    \includegraphics[width=.2\textwidth]{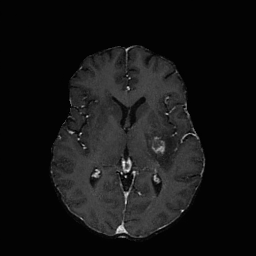}&%
    \includegraphics[width=.2\textwidth]{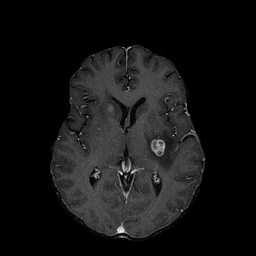}\\
    [-1pt]
    \includegraphics[width=.2\textwidth]{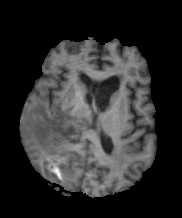}&%
    \includegraphics[width=.2\textwidth]{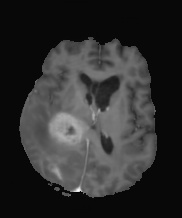}&%
    \includegraphics[width=.2\textwidth]{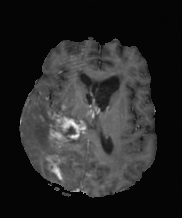}&%
    \includegraphics[width=.2\textwidth]{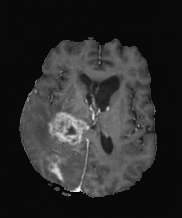}&%
    \includegraphics[width=.2\textwidth]{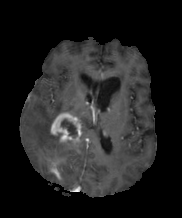}\\
    [-1pt]
    \includegraphics[width=.2\textwidth]{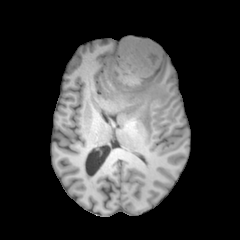}&%
    \includegraphics[width=.2\textwidth]{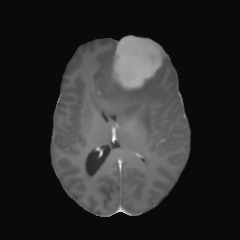}&%
    \includegraphics[width=.2\textwidth]{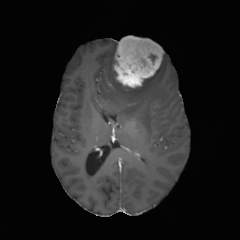}&%
    \includegraphics[width=.2\textwidth]{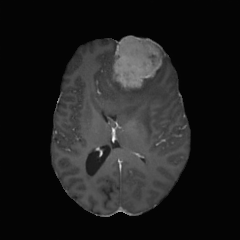}&%
    \includegraphics[width=.2\textwidth]{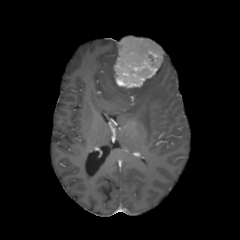}\\
    [2pt]
     \shortstack[t]{non-CE T1w\\(Input)}&
     \shortstack{Residual U-Net\\(PM)}&
     \shortstack[t]{RF\\Baseline}&
     \shortstack{\textbf{PMRF}\\(ours)}&
     \shortstack{CE T1w\\(GT)}
   \end{tabularx}
  }
  \caption{
    Qualitative comparison on three representative BraTS cases from the test set. \textbf{Top row:} Example 1 (BraTS-MET-00734-000). \textbf{Middle row:} Example 2 (BraTS-GLI-02508-102). \textbf{Bottom row:} Example 3 (BraTS-MEN-00686-000).
    }
  \label{fig:qualitative_examples}
\end{figure}

We qualitatively evaluate the top-performing Residual U-Net (Posterior Mean), RF baseline, and PMRF models (with a patch-overlap of 32 for all and $K=200$ integration steps for the RF baseline as well as the PMRF model) on the held-out test set.
For a set of representative cases, Figure~\ref{fig:qualitative_examples} compares the non-CE T1w input, Res.~U-Net (equivalent to PM), RF baseline, our two-stage PMRF, and the ground truth CE T1w image.
Example 1 depicts a metastases patient, example 2 a high-grade glioma patient, and example 3 a meningioma patient.
Across these test cases, the Res.~U-Net output accurately localizes lesions and captures their overall enhancement intensity but appears overly smooth, and muted in rim heterogeneity and fine texture.
The Stage~2 PMRF refinement consistently sharpens lesion boundaries, reinstates subtle vascular and margin details, and restores characteristic enhancement patterns, yielding synthetic images nearly indistinguishable from true post-contrast scans. However, Stage~2 can only refine what the posterior-mean predictor already suggests. If a subtle rim, micro‐metastasis, or vessel enhancement is entirely suppressed in Stage~1, Stage~2 has no signal to resurrect it; its perturbations stay within a narrow neighborhood of the Stage~1 output.
In contrast, the RF baseline directly predicts the CE signal from the perturbed non-CE image. However, this comes with the limitation of not predicting the CE strength faithful or missing subtle details as in examples 2 and 3.

\section{Conclusion}
We presented a two-stage PMRF framework for synthesizing CE T1w MRI from non-CE T1w inputs, explicitly targeting the perception-distortion trade-off in medical image synthesis.
By decoupling voxel-wise accuracy from perceptual refinement, our method achieves significant improvements in image realism (reducing axial FID by around 68.7\%), while maintaining acceptable distortion levels.
First quantitative and qualitative evaluations on a diverse dataset confirm that PMRF restores fine structural and textural details crucial for diagnosis.

\subsubsection*{Acknowledgements}
The authors acknowledge support from the DFG within the SPP2298 under project number 543939932 and from the Austrian Science Fund (FWF) project number 10.55776/COE12.

\appendix

\section{Experimental Setup and Hyperparameters}
\label{appendix:experiment_config}

The complete source code -- including data preprocessing scripts, model architectures, training routines, and evaluation tools -- is publicly available in our repository at \url{https://github.com/bbasti/pmrf-virtual-contrast}. Detailed, step-by-step instructions and comprehensive configuration files are also provided to ensure the full reproducibility of all experiments reported in this work.

\begin{figure}
  \centering
  \includegraphics[width=0.48\textwidth]{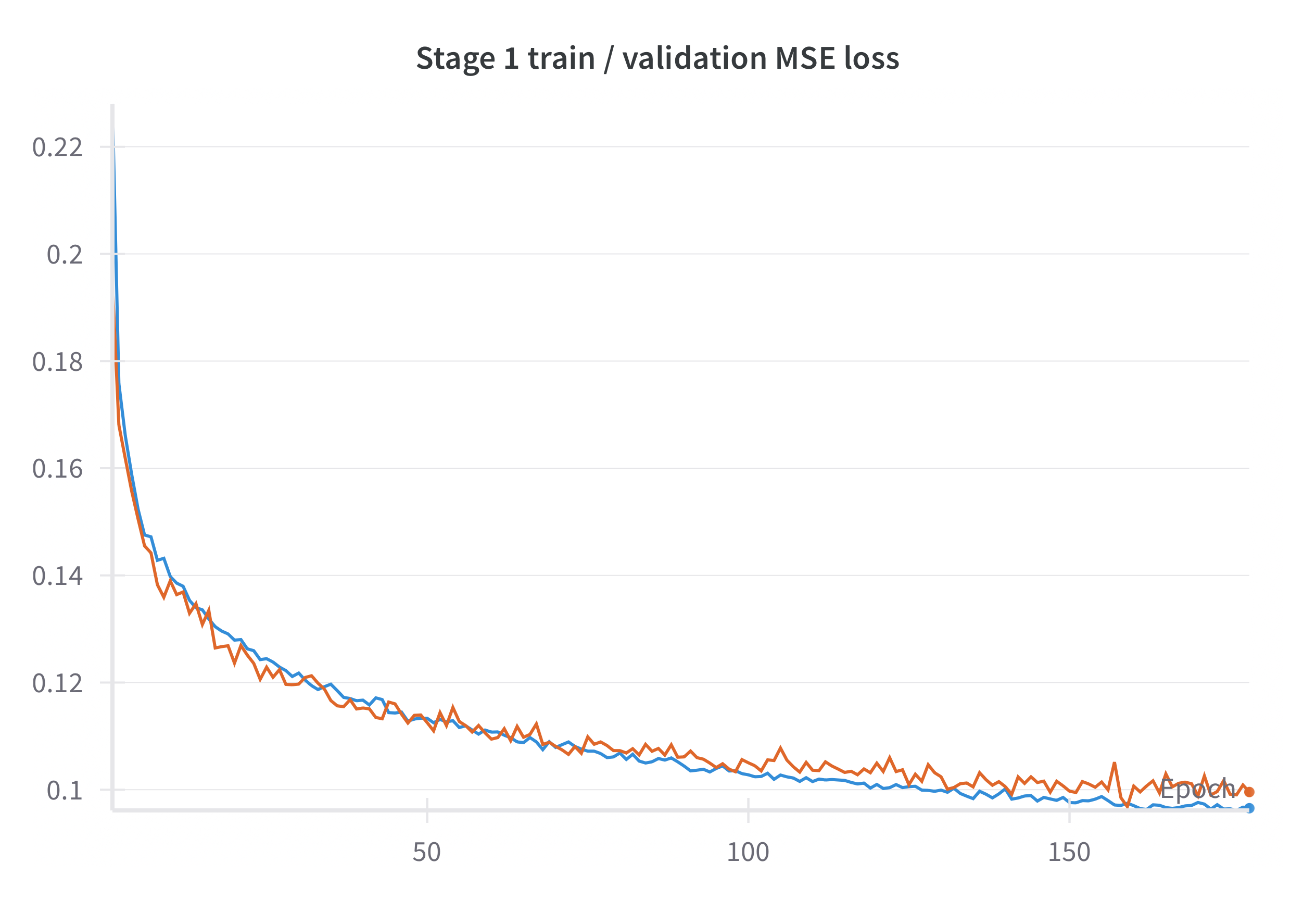}
  \includegraphics[width=0.48\textwidth]{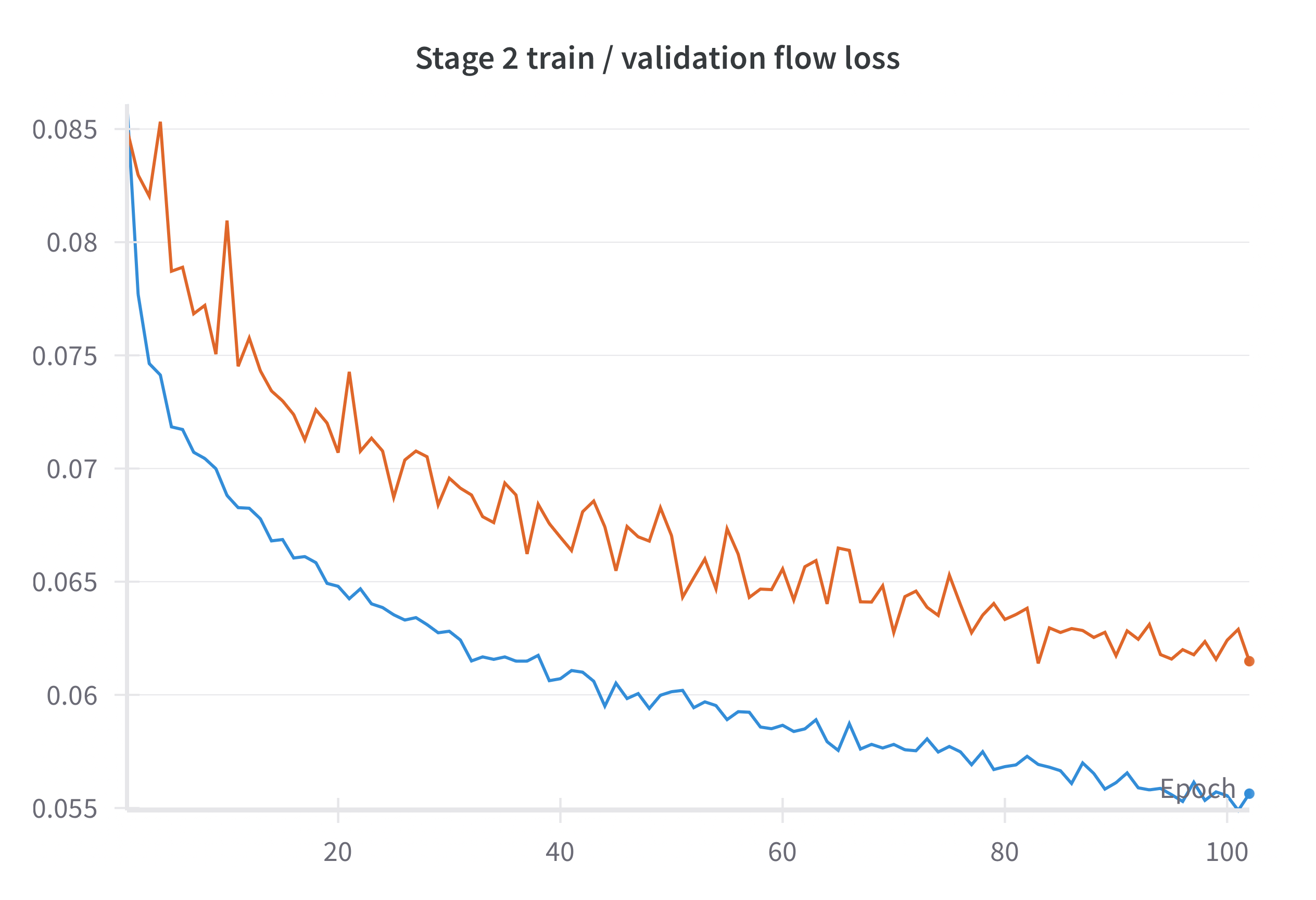}
  \caption{Stage~1 training (blue) and validation (orange) loss curves (left) and Stage~2 training (blue) and validation (orange) flow loss curves (right). The posterior-mean model converged steadily and was early-stopped at epoch 178 (wall-clock time: 1 d 8 h 13 m). The rectified-flow model reached its best validation loss at epoch 102 (1 d 4 h 44 m). All experiments were conducted on a single A100 80 GB GPU.}
  \label{fig:loss_curves_app}
\end{figure}

\begin{itemize}
  \item \textbf{Data splits:} 80\% for training, 15\% for validation (for early stopping), and 5\% held out as our test set.
  \item \textbf{Preprocessing:} 3D volumes were $Z$-score normalized, then padded in all dimensions to a minimum size of $64^3$ voxels. During training, we sampled eight random $64^3$-voxel patches per volume on the fly.
   \item \textbf{Training:} \begin{itemize}
  \item \textbf{Stage~1 (Posterior Mean):}
    \begin{itemize}
      \item Architecture: 3D residual U-Net
      \item Loss: Mean-squared error on $64^3$ patches.
      \item Optimizer: AdamW; initial learning rate $5\times10^{-4}$; cosine-annealing schedule over 200 epochs.
      \item Batch size: 128 patches.
      \item Early stopping: no improvement in validation loss for 20 epochs.
    \end{itemize}
  \item \textbf{Stage~2 (Rectified Flow) and RF baseline:}
    \begin{itemize}
      \item Architecture: 3D time-conditioned residual U-Net
      \item Loss: Flow-matching loss.
      \item Euler integration steps for rectified flow during training: 100.
      \item Optimizer, schedule, and early stopping: the same as Stage~1.
      \item Batch size: 64 patches.
    \end{itemize}
    \end{itemize}
  \item \textbf{Inference:}
    \begin{itemize}
      \item Patch size $64^3$, noise scale $\sigma_s=0.1$.
      \item Overlap: 4, 16 and 32 voxels (6.25\%, 25\%, 50\%)
      \item Euler integration steps in Stage~2 and RF baseline: \newline $K \in \{1,2,5,10,15,20,25,50,75,100,200\}$
    \end{itemize}
    \item \textbf{FID and KID computation:} We used the \texttt{torch-fidelity} package to compute Fréchet Inception Distance and Kernel Inception Distance. All 2D slices were resized to $512\times512$ pixels. Torch-fidelity extracts the full 2048-dimensional feature vectors from the \texttt{pool\_3} layer of a pretrained Inception-v3 model (PyTorch). 
\end{itemize}

\section{Full evaluation table}
\label{appendix:full_evaluation}

\begin{table}[H]
  \centering
  \caption{Full listing of all Stage~1 (PM), RF baseline and Stage~2 (PMRF) runs across overlaps and Euler steps on the 360-volume test set. Best overall values in each column are \textbf{bold}.}
{\scriptsize
\begin{tabular}{lcc|ccc|cc}
\hline
Model & Overlap & $K$ & MSE~$\downarrow$ & PSNR (dB)~$\uparrow$ & SSIM~$\uparrow$ & FID$_\text{axial}$~$\downarrow$ & KID$_\text{axial}$~$\downarrow$\\
\hline
\hline
PM (Res U-Net) & 4 & - & 0.0465 & 34.66 & 0.9516 & 40.57 & 0.0326 \\
PM (Res U-Net) & 16 & - & 0.0455 & 34.75 & 0.9521 & 40.31 & 0.0321 \\
PM (Res U-Net) & 32 & - & 0.0450 & \textbf{34.82} & \textbf{0.9524} & 39.76 & 0.0314 \\
\hline
RF Baseline & 32 & 1 & 0.0475 & 34.52 & 0.9506 & 39.75 & 0.0325 \\
RF Baseline & 32 & 2 & 0.0497 & 34.38 & 0.9508 & 28.20 & 0.0232 \\
RF Baseline & 32 & 5 & 0.0542 & 34.00 & 0.9483 & 18.23 & 0.0137 \\
RF Baseline & 32 & 10 & 0.0571 & 33.74 & 0.9465 & 14.62 & 0.0100 \\
RF Baseline & 32 & 15 & 0.0586 & 33.62 & 0.9456 & 13.56 & 0.0089 \\
RF Baseline & 32 & 20 & 0.0593 & 33.55 & 0.9451 & 13.10 & 0.0084 \\
RF Baseline & 32 & 25 & 0.0599 & 33.50 & 0.9448 & 12.60 & 0.0079 \\
RF Baseline & 32 & 50 & 0.0612 & 33.40 & 0.9441 & 12.2 & 0.0074 \\
RF Baseline & 32 & 75 & 0.0616 & 33.36 & 0.9438 & 11.86 & 0.0070 \\
RF Baseline & 32 & 100 & 0.0619 & 33.34 & 0.9437 & 11.85 & 0.0070 \\
RF Baseline & 32 & 200 & 0.0623 & 33.30 & 0.9434 & \textbf{11.66} & \textbf{0.0068} \\
\hline
PMRF (ours) & 4 & 1 & 0.0467 & 34.60 & 0.9498 & 39.03 & 0.0316 \\
PMRF (ours) & 4 & 2 & 0.0484 & 34.45 & 0.9498 & 28.91 & 0.0231 \\
PMRF (ours) & 4 & 5 & 0.0523 & 34.07 & 0.9470 & 20.42 & 0.0150 \\
PMRF (ours) & 4 & 10 & 0.0555 & 33.76 & 0.9441 & 17.61 & 0.0122 \\
PMRF (ours) & 4 & 15 & 0.0572 & 33.61 & 0.9425 & 17.65 & 0.0122 \\
PMRF (ours) & 4 & 20 & 0.0583 & 33.51 & 0.9415 & 17.81 & 0.0123 \\
PMRF (ours) & 4 & 25 & 0.0590 & 33.46 & 0.9410 & 17.98 & 0.0125 \\
PMRF (ours) & 4 & 50 & 0.0606 & 33.31 & 0.9395 & 18.57 & 0.0131 \\
PMRF (ours) & 4 & 75 & 0.0613 & 33.26 & 0.9390 & 18.91 & 0.0134 \\
PMRF (ours) & 4 & 100 & 0.0617 & 33.23 & 0.9387 & 19.05 & 0.0136 \\
PMRF (ours) & 4 & 200 & 0.0623 & 33.2 & 0.9382 & 19.37 & 0.0139 \\
PMRF (ours) & 16 & 1 & 0.0456 & 34.72 & 0.9507 & 39.18 & 0.0316 \\
PMRF (ours) & 16 & 2 & 0.0471 & 34.58 & 0.9508 & 29.17 & 0.0232 \\
PMRF (ours) & 16 & 5 & 0.0507 & 34.23 & 0.9484 & 19.96 & 0.0146 \\
PMRF (ours) & 16 & 10 & 0.0536 & 33.94 & 0.9458 & 16.36 & 0.0111 \\
PMRF (ours) & 16 & 15 & 0.0552 & 33.79 & 0.9444 & 15.93 & 0.0104 \\
PMRF (ours) & 16 & 20 & 0.0561 & 33.70 & 0.9436 & 15.87 & 0.0104 \\
PMRF (ours) & 16 & 25 & 0.0568 & 33.65 & 0.9430 & 16.01 & 0.0106 \\
PMRF (ours) & 16 & 50 & 0.0584 & 33.51 & 0.9417 & 16.18 & 0.0106 \\
PMRF (ours) & 16 & 75 & 0.0590 & 33.45 & 0.9412 & 16.24 & 0.0107 \\
PMRF (ours) & 16 & 100 & 0.0593 & 33.43 & 0.9410 & 16.34 & 0.0108 \\
PMRF (ours) & 16 & 200 & 0.0598 & 33.39 & 0.9405 & 16.67 & 0.0110 \\
PMRF (ours) & 32 & 1 & \textbf{0.0449} & 34.81 & 0.9516 & 40.41 & 0.0328 \\
PMRF (ours) & 32 & 2 & 0.0462 & 34.69 & 0.9518 & 29.55 & 0.0234 \\
PMRF (ours) & 32 & 5 & 0.0492 & 34.39 & 0.9501 & 19.66 & 0.0142 \\
PMRF (ours) & 32 & 10 & 0.0517 & 34.13 & 0.9482 & 15.03 & 0.0098 \\
PMRF (ours) & 32 & 15 & 0.0531 & 34.00 & 0.9472 & 14.12 & 0.0090 \\
PMRF (ours) & 32 & 20 & 0.0539 & 33.92 & 0.9465 & 13.60 & 0.0084 \\
PMRF (ours) & 32 & 25 & 0.0544 & 33.86 & 0.9461 & 13.27 & 0.0080 \\
PMRF (ours) & 32 & 50 & 0.0558 & 33.74 & 0.9451 & 12.71 & 0.0074 \\
PMRF (ours) & 32 & 75 & 0.0563 & 33.70 & 0.9447 & 12.51 & 0.0072 \\
PMRF (ours) & 32 & 100 & 0.0566 & 33.67 & 0.9445 & 12.65 & 0.0074 \\
PMRF (ours) & 32 & 200 & 0.0570 & 33.64 & 0.9442 & 12.46 & 0.0071 \\
\hline
\end{tabular}
}
  \label{tab:metrics_full_summary}
\end{table}

\bibliography{bibliography}

\end{document}